\documentclass[journal]{IEEEtran}
\IEEEoverridecommandlockouts

\usepackage{graphicx}
\usepackage{amsmath}
\usepackage{amssymb}
\usepackage{booktabs}
\usepackage{times}
\usepackage{epsfig}
\usepackage{mathtools}
\usepackage{bbm}
\usepackage{threeparttable}
\usepackage{multirow}
\usepackage{algorithm}
\usepackage{balance}
\usepackage[noend]{algpseudocode}
\usepackage{subcaption}
\usepackage{enumitem}
\usepackage{pifont}
\newcommand{\cmark}{\ding{51}}%
\newcommand{\xmark}{\ding{55}}%

\newcommand{\proposename}{Range-Aware Attention Network}
\newcommand{\proposenameshort}{RAANet}

\usepackage[pagebackref,breaklinks,colorlinks]{hyperref}

\def\BibTeX{{\rm B\kern-.05em{\sc i\kern-.025em b}\kern-.08em
    T\kern-.1667em\lower.7ex\hbox{E}\kern-.125emX}}

\begin{document}

\title{\proposename\ for LiDAR-based 3D Object Detection with Auxiliary Point Density Level Estimation}

\author{
Yantao Lu$^{1}$,~\IEEEmembership{} Xuetao Hao$^{2}$,~\IEEEmembership{} Shiqi Sun$^3$,~\IEEEmembership{} Weiheng Chai$^1$,~\IEEEmembership{} Senem Velipasalar$^1$,~\IEEEmembership{Senior~Member,~IEEE}\\
\thanks{
Authors are with $^1$Syracuse University, Syracuse, NY 13244 USA, $^2$University of Southern California, Los Angeles, CA 90089 USA, and $^3$Duke University, Durham, NC 27705, USA.

e-mail: ylu25@syr.edu, xuetaoha@usc.edu, shiqi.sun@duke.edu, wchai01@syr.edu, svelipas@syr.edu.
}
}

\maketitle

\begin{abstract}
3D object detection from LiDAR data for autonomous driving has been making remarkable strides in recent years. Among the state-of-the-art methodologies, encoding point clouds into a bird's eye view (BEV) has been demonstrated to be both effective and efficient. Different from perspective views, BEV preserves rich spatial and distance information between objects. Yet, while farther objects of the same type do not appear smaller in the BEV, they contain sparser point cloud features. This fact weakens BEV feature extraction using shared-weight convolutional neural networks (CNNs). In order to address this challenge, we propose \proposename\ (\proposenameshort), which extracts effective BEV features and generates superior 3D object detection outputs. The range-aware attention (RAA) convolutions significantly improve feature extraction for near as well as far objects. Moreover, we propose a novel auxiliary loss for point density estimation to further enhance the detection accuracy of \proposenameshort\ for occluded objects. It is worth to note that our proposed RAA convolution is lightweight and compatible to be integrated into any CNN architecture used for detection from a BEV. Extensive experiments on the nuScenes and KITTI datasets demonstrate that our proposed approach outperforms the state-of-the-art methods for LiDAR-based 3D object detection, with real-time inference speed of 16 Hz for the full version and 22 Hz for the lite version tested on nuScenes lidar frames. The code is publicly available at our Github repository \href{https://github.com/erbloo/RAAN}{https://github.com/erbloo/RAAN}.
\end{abstract}


\section{Introduction}
\label{sec:intro}

\begin{figure}
\centering
\includegraphics[width=0.45\textwidth]{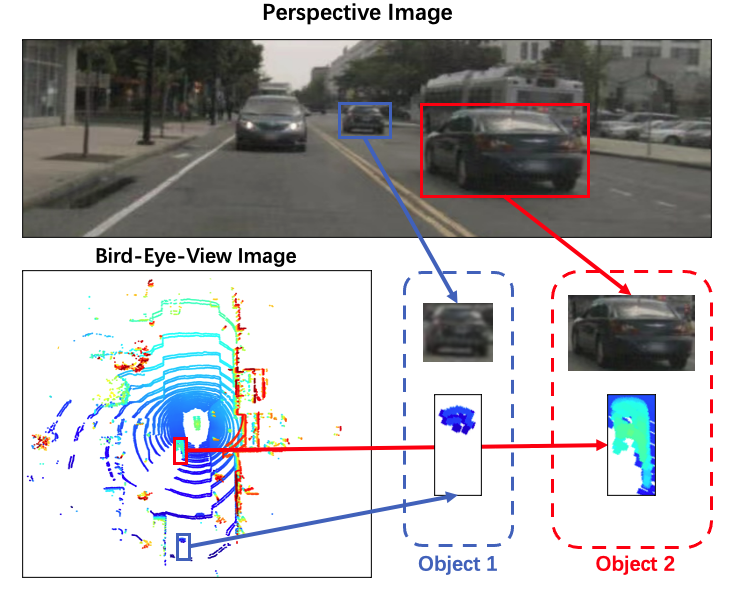}
\caption{\small \textbf{Motivation.} Different from camera images with perspective distortion, in BEV, objects that are farther from the ego-vehicle do not appear in smaller sizes, but contain sparser set of LiDAR points. Thus, an object in BEV may appear different depending on the distance to the ego-vehicle, which weakens BEV feature extraction using shared-weight CNNs.}
\label{fig:intro:compare}
\end{figure}

With the rapid improvement of processing units, and thanks to the success of deep neural networks, the perception of autonomous driving has been flourishing in recent years. 3D object detection from LiDAR sensors is one of the important capabilities for autonomous driving. Early works employ 3D convolutional neural networks (CNNs), which have slow processing speeds and large memory requirements. In order to decrease the memory requirements and provide real-time processing, recent methodologies leverage voxelization and bird's eye view (BEV) projection. Voxelization is widely implemented as a preprocessing method for 3D point clouds because of the computing efficiency, provided by more structured data, and performance accuracy~\cite{lang2019pointpillars}\cite{zhou2018voxelnet}\cite{zhou2020endtoend-multiview-fusion}\cite{xuetao2019segmentation}\cite{xu2021opencda}. In general, voxelization divides a point cloud into an evenly spaced grid of voxels, and then assigns 3D LiDAR points to their respective voxels. The output space preserves the Euclidean distance between objects and avoids overlapping of bounding boxes. This fact keeps object size variation in a relatively small range, regardless of their distance from the LiDAR, which benefits the shape regression during training. However, as shown in Figure~\ref{fig:intro:compare}, voxels that are farther away from the ego-vehicle contain significantly fewer LiDAR points than the near ones. This leads to a situation where different representations may be extracted for an object at different distances to the ego-vehicle. In contrast to perspective view, where feature elements are location-agnostic, BEV feature maps are location sensitive. Thus, different convolutional kernel weights should be applied to different locations of the feature map. In other words, location information should be introduced to the feature maps, and the convolutional kernels should be adjustable to the location information of corresponding feature maps. 

In this paper, we propose the \proposename\ (\proposenameshort), which contains the novel Range-Aware Attention Convolutional layer (RAAConv) designed for object detection from LiDAR BEV. RAAConv is composed of two independent convolutional branches and attention maps, which are sensitive to the location information of the input feature map. Our approach is inspired by the properties of BEV images, which are illustrated in Fig.~\ref{fig:intro:compare}. Points get sparser as the distance between an object and ego-vehicle increases. Ideally, for BEV feature maps, elements at different locations should be processed by different convolution kernels. However, applying different kernels will significantly increase the computational expense. In order to utilize the location information during BEV feature extraction, while avoiding heavy computation, we regard a BEV feature map as a composition of sparse features and dense features. We apply \textit{two different convolution kernels to simultaneously extract sparse and dense features}. Each extracted feature map has half the channel size of the final output. Meanwhile, range and position encodings are generated based on the input shape. Then, each range-aware attention heatmap is computed from the corresponding feature map and the range and position encodings. Finally, the attention heatmaps are applied on the feature maps to enhance feature representation. The feature maps generated from two branches are concatenated channel-wise as the RAAConv output. Details are presented in Sec.~\ref{ssec:mtd:attention_conv}.

In addition, effects of occlusion cannot be ignored, since the same object may have different point distributions under different amounts of occlusion. Thus, we propose an efficient auxiliary branch, referred to as the Auxiliary Density Level Estimation Module (ADLE), allowing \proposenameshort\ to take occlusion into consideration. Since annotating various occlusions is a time consuming and expensive task, we design the ADLE to estimate a point density level for each object. If there is no occlusion, point density levels for near objects are higher than those of far objects. However, if a near object is occluded, its point density level decreases. Therefore, by combining range information and density level information, we are able to estimate existence of the occlusion information. ADLE is only used in the training stage for providing density information guidance, and can be removed in inference state for computational efficiency.


\textbf{Contributions.} The main contributions of this work include the following:
\begin{itemize}[leftmargin=*]
\item We propose the RAAConv layer, which allows LiDAR-based detectors to extract more representative BEV features. In addition, RAAConv layer can be integrated into any CNN architecture used for LiDAR BEV.
\item We propose a novel auxiliary loss for point density estimation to help the main network learn occlusion related features. This proposed density level estimator further enhances the detection accuracy of \proposenameshort\ on occluded objects as well.
\item We propose the \proposename\ (\proposenameshort), which integrates the aforementioned RAA and ADLE modules. \proposenameshort\ is further optimized by generating anisotropic Gaussian heatmap based on the ground truth, which is introduced in Sec.~\ref{ssec:mtd:gaussian2d}. 
\item The code is available at the following GitHub repo~\cite{our-code}.
\end{itemize}

\section{Related Work}
\label{sec:related}

Majority of the object detection works can be categorized into two main groups: object detection with anchors and without anchors. Additionally, there exist works \cite{yang2018pixor}\cite{zhou2018voxelnet}\cite{lang2019pointpillars} that encode point cloud data in the early stage, but they are out of scope for object detection network refactorization.

\subsection{Object detection with anchors}
The fixed-shaped anchor regression approach \cite{girshick2014rich}\cite{girshick2015fast} \cite{ren2015faster} has been proposed such that intermediate features can be extracted. Girschick et al.~\cite{girshick2014rich} perform independent forward propagation on each region proposal. Region proposal task and object detection task are compounded in \cite{girshick2015fast}, which avoids heavy convolution operation overhead and preserves the precision throughout the network. To decrease the computation further, Ren et al.~\cite{ren2015faster} introduce a separate network to predict region proposals.
This removes the time-consuming selective search algorithm and lets the network learn the region proposals. These works employ the two-stage detection scheme. However, the region-proposal-joint process (e.g. RPN) is still slow and memory intensive, and thus not compatible with real-time detection. To reduce computation as well as memory footprint during both training and inference phases, one stage detectors are exploited in many works~\cite{liu2016ssd}\cite{redmon2016you}\cite{lin2017focal}, which can train object classification and bounding box regression directly, without pre-generated region proposals during training, and maintain a lower memory footprint at the same time. Redmon et al.~\cite{redmon2016you} redefine object detection as a single regression problem, which employs an end-to-end neural network to do a single forward propagation to detect objects. Liu et al.~\cite{liu2016ssd} develop a multi-resolution-anchor technique to detect objects of a mixture of a scale as well as learn the offset to a certain extent than learning the anchor. Lin et al.~\cite{lin2017focal} present a focal loss to tackle the dense and small object detection problem while dealing with class imbalance and inconsistencies. Zhou and Tuzel~\cite{zhou2018voxelnet} and Lang et al.~\cite{lang2019pointpillars} propose neural networks for point clouds, which open up new possibilities for 3D detection tasks. It should be noted that, although the aforementioned studies are real-time detection works, they actually involve the computation of all anchors' regression corresponding for different anchor scales.  

\subsection{Object detection without anchors}
To address the computation overhead and hyperparameter redundancy introduced by anchor regression, and effectively process point cloud encodings, anchor-free object detection has been exploited in many works \cite{tian2019fcos}\cite{zhang2021dardet}\cite{huang2015densebox}\cite{zhu2019feature}\cite{law2018cornernet}\cite{duan2019centernet}\cite{zhou2019bottom}.
Anchor-free object detection can be classified into two broad categories, namely center-based and key point-based approaches.

\begin{figure*}[t]
\centering
\includegraphics[width=\textwidth]{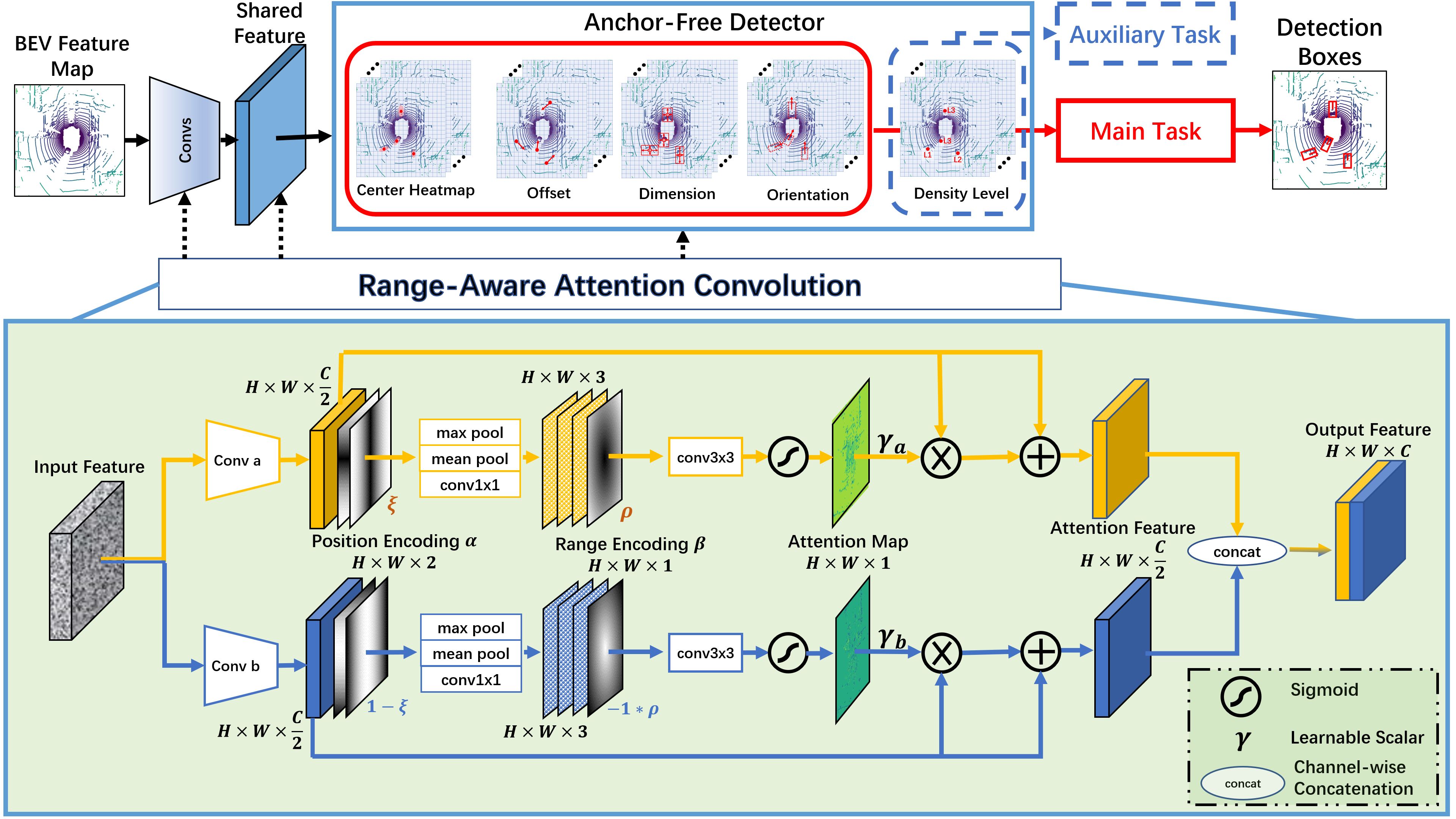}
\caption{\small \textbf{Overview of the proposed network.} The RPN-Points~\cite{lang2019pointpillars} is designed to extract the shared features from the BEV feature maps. The main detection heads and an auxiliary density estimation head are appended. The auxiliary head only exists during training and is omitted during inference. Finally, 3D bounding boxes are obtained from the main task. Our proposed RAAConv layer, illustrated in the lower part of the figure, replaces the original Conv2D layer. This layer contains two branches, and each branch applies position and range encodings to generate its range-aware attention heatmap. Then, extracted features in each branch are enhanced by the heatmap with a residual architecture. Those enhanced features are concatenated for the final output features.}
\label{fig:mtd:flow}
\end{figure*}

\subsubsection{Center-based approach}
In this approach~\cite{zhou2019centernet}\cite{yin2021center-point}, the center points of an object are used to define positive and negative samples, instead of IoU~\cite{tian2019fcos}\cite{huang2015densebox}\cite{zhu2019feature}. This method significantly reduces the computation cost by predicting four distances from positive samples to boundary of objects to generate a bounding box. Tian et al.~\cite{tian2019fcos} predict one point at each position instead of defining several bounding boxes with different scales. Huang et al.~\cite{huang2015densebox} propose the DenseBox, a unified fully convolutional neural network, which simultaneously detects multiple objects with confidence scores. This method is primarily applied to face detection and cars detection. Zhu et al.~ \cite{zhu2019feature} propose to attach online feature selection to each feature pyramid during training by setting up supervision signals through anchor-free branches. This work is demonstrated with feature pyramids of single-shot detectors. 

\subsubsection{Keypoint-based approach}
In this approach~\cite{law2018cornernet}\cite{duan2019centernet}\cite{zhou2019bottom}, the key points are located by several predefined methods or self-learned models, and then generate the bounding boxes to classify the objects. Law and Deng~\cite{law2018cornernet} propose corner pooling, which aggregates the features of objects on the corner and groups corners with associative embeddings. Zhou et al.~\cite{zhou2019bottom} develop a framework to predict four corner coordinates and center points simultaneously. However, the grouping algorithm and key points selection in these works introduce redundant false positive samples and effect the performance of detectors. To address this problem, Duan et al.~\cite{duan2019centernet} use both center points and corner points, which directly replace associative embedding with centripetal shifts. Although these paradigms are mainly proposed for optimizing data annotations, we have found that it also plays an essential role in feature representation and detection accuracy. In order to extract representative features, we focus on two major components: \textit{range-aware feature extraction} and \textit{occlusion supervision}. 
\section{\proposename}
\label{sec:mtd}

\subsection{Overview of the \proposenameshort}
\label{ssec:mtd:overview}
The main architecture of our proposed \proposename\ (\proposenameshort) is presented in Fig.~\ref{fig:mtd:flow}.~We incorporate ideas from CenterNet~\cite{zhou2019centernet}\cite{yin2021center-point} to build an anchor-free detector, and also introduce two novel modules: the Range-Aware Attention Convolutional Layer (RAAConv) and the Auxiliary Density Level Estimation Module (ADLE).

\proposenameshort\ takes the 3D LiDAR points as input and generates a set of 3D oriented bounding boxes as the output. Inspired by VoxelNet~\cite{zhou2018voxelnet}, we implement a feature extractor with 3D sparse convolutional network on the voxelized point clouds. This resulting features are reshaped and formed into BEV feature maps, which can be regarded as a multi-channel 2D image. The Region Proposal Network (RPN) takes this BEV feature maps as input and employs multiple down-sample and up-sample blocks to generate a high-dimensional feature map. In addition to the detection heads in main task, we propose an auxiliary task for point density level estimation to achieve better detection performance. Meanwhile, the RAAConv layers are utilized in all convolutional modules. The details of the RAAConv layer are shown in the bottom half of Fig.~\ref{fig:mtd:flow}. Center probability heatmap is generated from the detector head, and maximums in local neighborhoods are treated as the centers of ground-truth bounding boxes (bbox). The regression of bbox attributes, including offsets, dimensions and orientations, are computed from the other heads in the main detection task. Parallel to the main task, the proposed ADLE, as an auxiliary task, estimates the point density level for each bounding box. In general, ADLE classifies each bounding box into different density levels, based on the number of points in each bbox. ADLE is a portable block for the whole network, i.e. it is employed only during training and is not used during inference for higher computational efficiency.

For the training phase, the loss components, namely the heatmap classification loss, bbox regression loss and density-level auxiliary loss are denoted as $\mathit{l}_{hm}$, $\mathit{l}_{box}$ and $\mathit{l}_{aux}$, respectively. The total loss function of \proposenameshort\ is formulated as
\begin{equation}
{\small
\mathit{l} = \mathit{l}_{hm} + \lambda_{box} \mathit{l}_{box} + \lambda_{aux} \mathit{l}_{aux},
}
\label{eq:mtd:overall_loss}
\end{equation}
where $\lambda_{box}$ and $\lambda_{aux}$ are scalar weights for balancing the multi-task loss. $l_{hm}$ and $l_{aux}$ are designed as penalty-reduced focal losses~\cite{lin2017focalloss}\cite{zhou2019centernet}, and $l_{box}$ is designed as the smooth L1 loss.

\subsection{Range-Aware Attentional Convolutional Layer}
\label{ssec:mtd:attention_conv}
We propose the Range-Aware Attentional Convolutional Layer (RAAConv), which, by leveraging the specially designed attention blocks, becomes sensitive to range and position information. This is the major difference between the proposed RAAConv and a traditional convolutional layer. RAAConv is employed in the aforementioned RPN and detection heads.~As shown in Fig.~\ref{fig:mtd:flow}, RAAConv first utilizes two sets of convolutional kernels to extract an intermediate feature map for each branch. Then, the position encodings are generated and embedded into intermediate feature maps. Two range-aware heatmaps are calculated by a series of convolution and pooling operations using the intermediate feature maps and range encodings. More specifically, given an input feature map $I\in \mathbb{R}^{H_{in}\times W_{in}\times C_{in}}$, two separate intermediate feature maps, $\mathbf{F_a}, \mathbf{F_b}\in \mathbb{R}^{ H_{out}\times W_{out}\times \frac{C_{out}}{2}}$, are generated by
\begin{equation}
{\small
\begin{aligned}
\mathbf{F_i} = \zeta(\mathbf{I}, \mathbf{\mathit{w_F^i}}), i\in {a, b},
\end{aligned}
}
\label{eq:mtd:conv_ab}
\end{equation}
where $\zeta(\cdot)$ represents the convolution operation and $\mathbf{w_F^i}$ denotes the corresponding convolution kernel.

Meanwhile, the position encoding map $\mathbf{\xi}$ $\in$ $\mathbb{R}^{H_{out}\times W_{out}\times2}$ and range encoding map $\mathbf{\rho}$ $\in \mathbb{R}^{H_{out}\times W_{out}\times1}$ are generated from the intermediate feature map $\mathbf{F_a}$. $\mathbf{\xi}$ and $\mathbf{\rho}$ are calculated from shape information and do not depend on the specific values inside the tensor $F_a$. There are two components of $\mathbf{\xi}$: row encoding $\mathit{r}$ and column encoding $\mathit{c}$. $\mathbf{\xi}$ and $\mathbf{\rho}$ are generated as follows:
\begin{equation}
{\small
\begin{aligned}
\mathit{r_{ij}}=&\frac{2|i - \frac{H}{2}|}{H}, \,\,\, \mathit{c_{ij}}=\frac{2|j - \frac{H}{2}|}{W} \\
&\mathit{\rho_{ij}}=2\sqrt{\mathit{r_{ij}}^2+\mathit{c_{ij}}^2}-1 \\
\mathit{s.t.}\ i\in &\{1,2,...,H\},\ j\in \{1,2,...,W\}
\end{aligned}
}
\label{eq:mtd:pos_encode}
\end{equation}
Values of $\mathit{r, c}$ are bounded in $[0, 1]$. Values of $\mathit{\rho}$ are bounded in $[-1,1]$.

Since the proposed network is designed by using two different convolution kernels, to extract dense and sparse features separately, we append $\mathbf{\xi}$ to intermediate feature map $\mathbf{F_a}$ channel-wise, and reversely append $1-\mathbf{\xi}$ to $\mathbf{F_b}$. The generated features are denoted by $\mathbf{F_a'}$ and $\mathbf{F_b'}$. Maximum pooling, mean pooling and $1\times1$ conv2D are applied on $\mathbf{F_a'}$ and $\mathbf{F_b'}$ separately to obtain the spatial embeddings $\mathbf{F_a^\xi}$ and $\mathbf{F_b^\xi}$. Then, similar to positional appending, the range encoding map $\mathbf{\rho}$ is appended to $\mathbf{F_a^\xi}$, and $-\mathbf{\rho}$ is appended to $\mathbf{F_b^\xi}$. The appended features are processed by a $3\times3$ conv2D layer followed by sigmoid activation to obtain the range-aware attention heatmaps $\mathbf{f_a}$ and $\mathbf{f_b}$. The heatmaps $\mathbf{f_a}$ and $\mathbf{f_b}$ are then multiplied by learnable scalars $\gamma_a$ and $\gamma_b$, respectively. $\gamma_a$ and $\gamma_b$ are initialized as $1.0$ and their values are gradually learned 
during training~\cite{zhang2019self}. The output feature maps $\mathbf{F_a''}$ and $\mathbf{F_b''}$ are calculated from $\mathbf{F_a'}, \mathbf{f_a}, \gamma_a$, and $\mathbf{F_b'}, \mathbf{f_b}, \gamma_b$ using a residual connection, which is defined as:
\begin{equation}
\small
\begin{aligned}
\mathbf{F_i''}=(\gamma_i \mathbf{f_i}) \otimes \mathbf{F_i'} + \mathbf{F_i'}, \,\,\,\,\,\,i\in {a, b}
\end{aligned}
\label{eq:mtd:feature_output_branch}
\end{equation}
where $\otimes$ is the operation performing channel-wise stacking and element-wise multiplication. The final output for RAAConv is the channel-wise concatenation of $\mathbf{F_a''}$ and $\mathbf{F_b''}$, which has a size of $H_{out}\times W_{out}\times C_{out}$.


\begin{figure}
\centering
\includegraphics[width=0.45\textwidth]{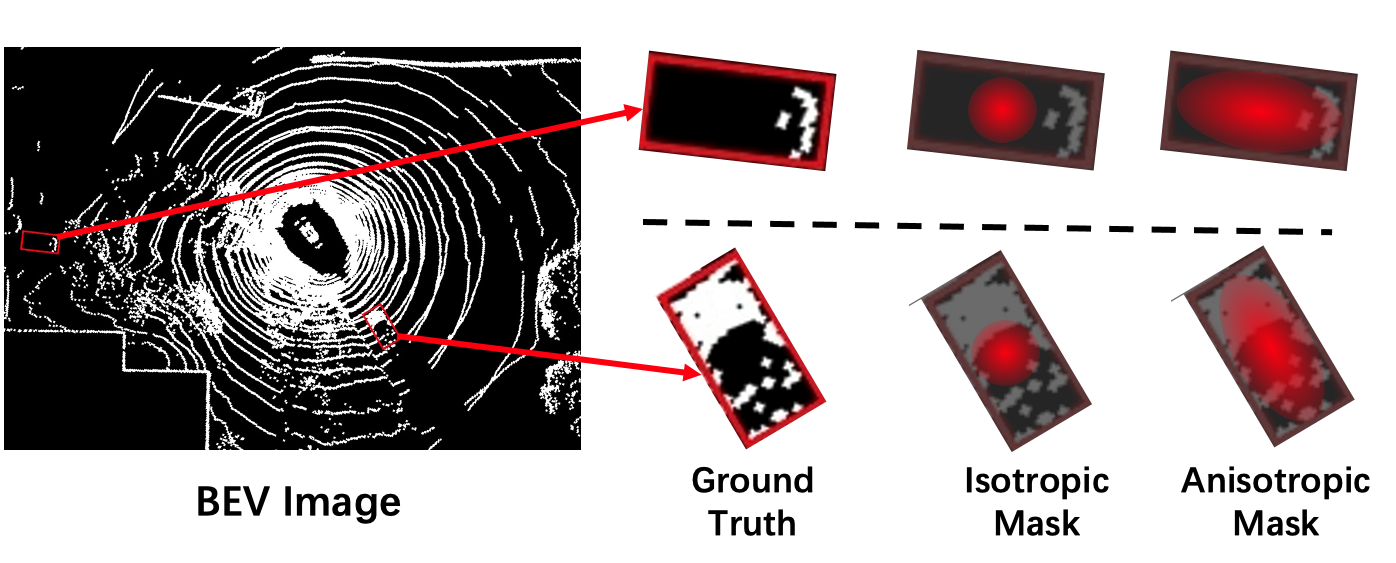}
\caption{\small 
\textbf{Illustration of Anisotropic Gaussian Mask.} Traditional Isotropic Gaussian Mask neglects the rotation of the bounding box. We propose an Anisotropic Gaussian Mask to better fit to the ground truth. \textit{The Anisotropic mask is able to introduce the heading information, which is beneficial to bbox regression as well.}
}
\label{fig:intro:gaussian}
\end{figure}

It is worth noting that the proposed RAAConv can be readily plugged in any convolutional network for LiDAR BEV detection.

\subsection{Auxiliary Density Level Estimation Module (ADLE)}
\label{ssec:mtd:density_level_estimation}
The Auxiliary Density Level Estimation Module (ADLE) is an additional classification head that is parallel to the main task heads. The ADLE module indicates the point density level for each bbox. In general, we design the ADLE module to estimate the number of LiDAR points inside a detected bbox. We have empirically found out that this estimation is straightforward yet beneficial for the detector. Moreover, we divide the number of points into three density bins and let the ADLE perform a classification task, instead of a value regression task, which helps training to converge better. More specifically, the three bins are labeled as sparse (label $1$), adequate (label $2$) and dense (label $3$). We have statistically analyzed the distribution of the number of points and employed two thresholds $T_0^i, T_1^i$ to determine the density level for each object class $i$. The three density levels are defined as:

\begin{equation}
\small{
D_i(n)=\left\{
\begin{matrix}
\begin{aligned}
&1,\ \text{if}\ n < T_0^i \\ 
&2,\ \text{if}\ T_0 \leqslant n < T_1^i \\
&3,\ \text{if}\ n\geqslant T_1^i\\
\end{aligned}
\end{matrix}
\right\}
}
\label{eq:mtd:occlusion}
\end{equation}
where $n$ denotes the number of points, and $D_i$ represents the density level for the $i$-th class. As shown in Fig.~\ref{fig:intro:gaussian}, the unoccluded instances near the ego-vehicle have high point density levels and faraway instances have low density levels. However, when an instance is occluded, the density level will be lower than the unoccluded case at the same distance. Since we incorporate position and range information into RAAConv, ADLE is able to further help \proposenameshort\ extract occlusion information by supervising the point density levels.

\subsection{Anisotropic Gaussian Mask}
\label{ssec:mtd:gaussian2d}
Anisotropic Gaussian Mask represents the centerness probability for each position in an annotated bbox. The centerness heatmap is generated as the ground truth for the classification head. In this work, we propose an Anisotropic Gaussian Mask, shown in Fig.~\ref{fig:intro:gaussian}, to generate a 2D Gaussian distribution, to be used for the centerness probabilities of a given oriented bbox. More specifically, the Gaussian distribution $\textit{N}(\mu,\Sigma)$ is designed in an anisotropic way, i.e. the $\sigma$ values for the 2 dimensions, which are the diagonals of $\Sigma$, and denoted as $\sigma_0, \sigma_1$, are different, and their values are determined from the length and width of the bbox, respectively. The procedure for generating the center heatmap, denoted by $\mathbf{\mathfrak{M}}$, can be formulated as:
\begin{equation}
\begin{aligned}
\mathbf{\mathfrak{M}}_{C_H,C_W,H,W,s}(x,y)=\mathbbm{1}_{\mathbf{\mathfrak{B}}}(\mathbf{\emph{N}}_{\mathbf{\mu,\Sigma}}(x,y))
\end{aligned}
\label{eq:mtd:gaussian_mask}
\end{equation}
where,
\begin{equation}
{\small
\begin{aligned}
\mathbf{\mathfrak{B}}=\{ x\in[C_W-\frac{W}{2}, &C_W+\frac{W}{2}] \cap y\in[C_H-\frac{H}{2}, C_H+\frac{H}{2}]\} \\
&\mathbf{\mu}=[C_W,C_H]^T \\
&\mathbf{\Sigma}=\begin{bmatrix}
W/d_{cls} & 0\\ 
0 & H/d_{cls}
\end{bmatrix}
\end{aligned}
}
\label{eq:mtd:gaussian_mask_cont}
\end{equation}
and $(C_H,C_W)$ and $(H,W)$ denote the center location and dimensions of a bbox, respectively. $s$ indicates the ground truth class for the bbox. $\mathbbm{1}(\cdot)$ is the indicator function and $\mathbf{\mathfrak{B}}$ denotes the set of locations that are inside the bbox. $d_{cls}$ is a decay factor determining the sharpness of generated heatmap. The value of $d_{cls}$ depends on the class of a given bbox.

\subsection{Loss Functions} \label{ssec:mtd:loss_func}
Following Eq.~(\ref{eq:mtd:overall_loss}), the center heatmap loss $\mathit{L}_{hm}$ is defined as:
\begin{equation}
{\small
\begin{aligned}
\mathit{L}_{hm}=- &\frac{1}{N}\sum_{\tilde{p}}\{(1-h)^\alpha \textup{log}(h)I_{\hat{h}>1-\epsilon}+ \\
& (1-\hat{h})^\beta h^\alpha\textup{log}(1-h)I_{\hat{h}\leqslant 1-\epsilon} \},
\end{aligned}
}
\label{eq:mtd:center_loss}
\end{equation}
where $h$ and $\hat{h}$ are the ground-truth and predicted heatmap values, respectively. $\epsilon$ is a small value for numerical stability. $\alpha$ and $\beta$ are the parameters for controlling the penalty-reduced focal loss. For all the experiments, we follow the setting in \cite{sun2021rsn}, i.e. $\epsilon=1e-3$, $\alpha=2$ and $\beta=4$.

The bounding box head is responsible for the location, dimension and orientation regression, which are represented by the following parameters: $\mathbf{\Gamma}=[\delta_x, \delta_y, \delta_z, l, w, h, \theta]$, where $\delta_x, \delta_y, \delta_z$ are the box center offsets with respect to the corresponding voxel centers; and $l, w, h $ and $\theta$ denote the length, width, height and yaw of a bounding box. We apply smoothed L1 (SL1) losses to regress $sin(\theta)$ and $cos(\theta)$ for the $\theta$ attribute, and directly regress actual values for other attributes:
\begin{equation}
{\small
\begin{aligned}
\mathit{L}_{\theta_i}=& SL1(sin(\theta_i)-sin(\hat{\theta}_i)) + \\
& SL1(cos(\theta_i)-cos(\hat{\theta}_i))
\end{aligned}
}
\label{eq:mtd:bbox_yaw_loss}
\end{equation}
\begin{equation}
{\small
\begin{aligned}
\mathit{L}_{\mathbf{\Gamma_i}\setminus \theta}=& SL1(\mathbf{\Gamma_i}\setminus \theta-\mathbf{\hat{\Gamma}_i}\setminus \hat{\theta})
\end{aligned}
}
\label{eq:mtd:bbox_other_loss}
\end{equation}
where, $\hat{\cdot}$ denotes predicted values corresponding to original symbols. The total loss for box regression is:
\begin{equation}
{\small
\begin{aligned}
\mathit{L}_{box} = \frac{\mathbbm{1}_i}{N}\sum_{i}(\mathit{L}_{\theta_i} + \mathit{L}_{\mathbf{\Gamma_i}\setminus \theta}),
\end{aligned}
}
\label{eq:mtd:bbox_total_loss}
\end{equation}
where $N$ is the number of box samples and $\mathbbm{1}_i$ equals 1 if the ground truth heatmap value of the $i$th feature map pixel is larger than a threshold $\tau$, which is set to 0.2 in the experiments.

\begin{table*}[t]
\centering
\begin{tabular}{l|cccccccccc|c}
\hline
                                       & Car            & Truck          & Bus            & Trailer        & \begin{tabular}[c]{@{}c@{}}Cons.\\ Veh.\end{tabular} & Ped.     & M.cyc.     & Bicycle        & \begin{tabular}[c]{@{}c@{}}Traffic\\ Cone\end{tabular} & Barrier        & mAP$\uparrow$         \\ \hline
PointPillars~\cite{lang2019pointpillars} & 0.760          & 0.310          & 0.321          & 0.366          & 0.113                                                          & 0.640          & 0.342          & 0.140          & 0.456                                                  & 0.564          & 0.401          \\
SA-Det3D~\cite{bhattacharyya2021sa} & 0.812          & 0.438          & 0.572          & 0.478          & 0.113                                                          & 0.733          & 0.321          & 0.079          & 0.606                                                  & 0.553          & 0.470          \\
InfoFocus~\cite{wang2020infofocus} & 0.779          & 0.314          & 0.448          & 0.373          & 0.107                                                          & 0.634          & 0.290          & 0.061          & 0.465                                                  & 0.478          & 0.395          \\
CyliNetRG~\cite{rapoport2021cylinet} & 0.850          & 0.502          & 0.569          & 0.526          & 0.191                                                          & 0.843          & 0.586          & 0.298          & 0.791                                         & 0.690          & 0.585          \\
SARPNET~\cite{YE2020sarpnet}           & 0.599          & 0.187          & 0.194          & 0.180          & 0.116                                                          & 0.694          & 0.298          & 0.142          & 0.446                                                  & 0.383          & 0.324          \\
PanoNet3D~\cite{chen2020panonet3d}     & 0.801          & 0.454          & 0.540          & 0.517          & 0.151                                                          & 0.791          & 0.531          & 0.313          & 0.719                                                  & 0.629          & 0.545          \\
PolarStream~\cite{chen2021polarstream} & 0.809          & 0.381          & 0.471          & 0.414          & 0.195                                                          & 0.802          & 0.614          & 0.299          & 0.753                                                  & 0.640          & 0.538          \\
MEGVII~\cite{zhu2019MEGVII}            & 0.811          & 0.485          & 0.549          & 0.429          & 0.105                                                          & 0.801          & 0.515          & 0.223          & 0.709                                                  & 0.657          & 0.528          \\
PointRCNN\cite{shi2020pcdet} & 0.810          & 0.472          & 0.563          & 0.510          & 0.141                                                          & 0.766          & 0.422          & 0.134          & 0.667                                                  & 0.614          & 0.510          \\
SSN-v2~\cite{zhu2020ssn}               & 0.824          & 0.418          & 0.461          & 0.480          & 0.175                                                          & 0.756          & 0.489          & 0.246          & 0.601                                                  & 0.612          & 0.506          \\
ReconfigPP-v2~\cite{wang2020reconfigurable} & 0.758          & 0.272          & 0.395          & 0.380          & 0.065                                                          & 0.625          & 0.152          & 0.002          & 0.257                                                  & 0.349          & 0.325          \\
ReconfigPP-v3~\cite{wang2020reconfigurable}\cite{wang2021probabilistic} & 0.814          & 0.389          & 0.430          & 0.470          & 0.153                                                          & 0.724          & 0.449          & 0.226          & 0.583                                                  & 0.614          & 0.485          \\
HotSpotNet~\cite{chen2020hotspots} & 0.831          & 0.509          & 0.564          & 0.533          & 0.230                                                          & 0.813          & {\color{blue}0.635} & 0.366          & 0.730                                                  & 0.716          & 0.593          \\
MMDec3D~\cite{mmdet3d2020}       & 0.847          & 0.490          & 0.541          & 0.528          & 0.216                                                          & 0.793          & 0.560          & {\color{blue}0.387}          & 0.714                                                  & 0.674          & 0.575          \\
CenterPoint~\cite{yin2021center-point} & 0.852          & 0.535          & {\color{blue}0.636} & {\color{blue}0.560}          & 0.200                                                          & 0.846          & 0.595          & 0.307          & 0.784                                                  & 0.711          & 0.603          \\
CVCNet-ens~\cite{chen2020everyCVC}     & 0.826          & 0.495          & 0.594          & 0.511          & 0.162                                                          & 0.830          & 0.618          & \textbf{0.388}          & 0.697                                                  & 0.697          & 0.582          \\ \hline
\proposenameshort-lite (Ours)                         & {\color{blue}0.860} & {\color{blue}0.546} & {\color{blue}0.636}          & 0.553          & {\color{blue}0.237}                                                 & {\color{blue}0.856} & 0.633          & 0.340          & {\color{blue}0.794}                                                  & {\color{blue}0.749} & {\color{blue}0.620}          \\ 
\proposenameshort (Ours)                     & \textbf{0.864} & \textbf{0.572} & \textbf{0.648}          & \textbf{0.569}          & \textbf{0.240}                                                 & \textbf{0.857} & \textbf{0.641}          & 0.371          & \textbf{0.829}                                                  & \textbf{0.761} & \textbf{0.635} \\ \hline
\end{tabular}
\caption{\small 
\textbf{Comparison with the state-of-the-art models} for 3D detection on the nuScenes test dataset. We show AP scores for each class and overall mAP scores. The best results are marked in \textbf{bold} and second best results are marked in {\color{blue}{blue}} color. The \proposenameshort\ refers to the network that applies RAAConv layers throughout all modules. The \proposenameshort-lite refers to the network that applies RAAConv layers on the detection heads.
}
\label{tab:exp:main_results}
\end{table*}
\begin{table*}[]
\centering
\begin{tabular}{l|cccccc|c}
\hline
\multicolumn{1}{c|}{} & mAP$\uparrow$   & mATE$\downarrow$  & mASE$\downarrow$  & mAOE$\downarrow$  & mAVE$\downarrow$  & mAAE$\downarrow$  & NDS$\uparrow$   \\ \hline
PointPillars          & 0.401 & 0.392 & 0.269 & 0.476 & 0.270 & \textbf{0.102} & 0.550 \\
SA-Det3D              & 0.470 & 0.317 & 0.248 & 0.438 & 0.300 & 0.129 & 0.592 \\
InfoFocus             & 0.395 & 0.363 & 0.265 & 1.132 & 1.000 & 0.395 & 0.395 \\
CyliNetRG             & 0.585 & 0.272 & 0.243 & 0.383 & 0.293 & 0.126 & 0.661 \\
SARPNET               & 0.324 & 0.400 & 0.249 & 0.763 & 0.272 & 0.090 & 0.484 \\
PanoNet3D             & 0.545 & 0.298 & 0.247 & 0.393 & 0.338 & 0.136 & 0.631 \\
PolarStream           & 0.538 & 0.339 & 0.264 & 0.491 & 0.339 & 0.123 & 0.613 \\
MEGVII                & 0.528 & 0.300 & 0.247 & 0.379 & 0.245 & 0.140 & 0.633 \\
PointRCNN             & 0.510 & 0.301 & 0.245 & 0.378 & 0.326 & 0.138 & 0.616 \\
SSN-v2                & 0.506 & 0.339 & 0.245 & 0.429 & 0.266 & 0.087 & 0.616 \\
ReconfigPP-v2         & 0.325 & 0.388 & 0.250 & 0.486 & 0.306 & 0.137 & 0.506 \\
ReconfigPP-v3         & 0.485 & 0.328 & 0.244 & 0.439 & 0.274 & 0.244 & 0.590 \\
HotSpotNet            & 0.593 & 0.274 & \textbf{0.239} & 0.384 & 0.333 & 0.133 & 0.660 \\
MMDec3D               & 0.575 & 0.316 & 0.256 & 0.409 & 0.236 & {\color{blue}0.124} & 0.653 \\
CenterPoint           & 0.603 & {\color{blue}0.262} & \textbf{0.239} & {\color{blue}0.361} & 0.288 & 0.136 & 0.673 \\
CVCNet-ens            & 0.582 & 0.284 & 0.241 & 0.372 & 0.224 & 0.126 & 0.666 \\ \hline
RAANet-lite (Ours)    & {\color{blue}0.620} & 0.277 & 0.244 & 0.365 & {\color{blue}0.212} & 0.126 & {\color{blue}0.687} \\
RAANet (Ours)         & \textbf{0.635} & \textbf{0.259} & 0.240 & \textbf{0.326} & \textbf{0.210} & 0.125 & \textbf{0.702} \\ \hline
\end{tabular}
\caption{\small 
\textbf{Comparison with the state-of-the-art models} for 3D detection on the nuScenes test dataset for different performance metrics. The best results are marked in \textbf{bold} and second best results are marked in {\color{blue}{blue}} color.
}
\label{tab:exp:main_results_othermetrics}
\end{table*}

\section{Experiments}
\label{sec:exp}
We introduce the implementation details of the proposed \proposenameshort\, and evaluate its performance in multiple experiments. The inference results are uploaded to and evaluated by the nuScenes official benchmark~\cite{nusceneslink}.~The diagnostic analysis of each component is shown in the ablation studies in Sec.~\ref{sec:exp:ablation}.

\subsection{Datasets}
\label{ssec:exp:datasets_discription}
We perform the evaluations on the nuScenes~\cite{nuscenes2019} and KITTI~\cite{Geiger2012kitti} datasets, which are publicly available, large-scale and challenging datasets for autonomous driving. The full nuScenes dataset includes approximately 390K LiDAR sweeps and 1.4M object bounding boxes in 40K key frames. The dataset contains 1000 scenes in total, of which 700 are separated for training, 150 for validation and 150 for testing. There are a total of 10 classes: $\{$\emph{car, truck, bus, trailer, construction vehicle, traffic cone, barrier, motorcycle, bicycle, pedestrian}$\}$. The scenes of 20 sec. length are manually selected to cover a diverse and interesting set of driving maneuvers, traffic situations and unexpected behaviors. The dataset was captured with a 32-channel spinning LiDAR sensor with 360 degree horizontal and -30 to +10 degree vertical FOV. The sensor captures points, which are within 70 meters with $\pm2$ cm accuracy and returns up to 1.39M points per second. 

We also use the KITTI dataset to benchmark performance. The 3D object detection dataset consists of 7481 training and 7518 testing samples for the point clouds, including a total of 80256 labeled objects. Following the experimental setting used in~\cite{lang2019pointpillars}, we divide the official training samples into 3712 training samples and 3769 validation samples. Meanwhile, considering the test submission requirement, we create 784 samples as a minival validation set, and use the remaining 6697 samples to train. Since the KITTI benchmark only provides the ground truth annotations of objects that are visible in camera images, we followed the standard literature practice on KITTI of only using LiDAR points that project into the image, and trained one network for cars to benchmark our model performance.


\subsection{Implementation Details}
\label{ssec:exp:imp_detail}
\proposenameshort\ is implemented in PyTorch framework~\cite{paszke2017Pytorch} with an anchor-free object detection architecture. The weights $\lambda_{box}$ and $\lambda_{aux}$ in loss function (Eq.~\ref{eq:mtd:overall_loss}) are set to $0.25$ and $0.2$, respectively. The decay factor $d$ in Sec.~\ref{ssec:mtd:gaussian2d} is set to $3$ for large size classes (car, truck, bus, trailer, construction\_vehicle), to $6$ for small object classes (traffic\_cone, barrier, motorcycle, bicycle, pedestrian). As for the parameters in Sec.~\ref{ssec:mtd:loss_func}, we set $\epsilon$, $\alpha$, $\beta$ and $\tau$ to $0.001$, $2.0$, $4.0$ and $0.2$, respectively.

For nuScenes, all classes are trained end-to-end. The resolution for voxelization is set to $[0.075\text{m}, 0.075\text{m}, \text{0.2m}]$ and the voxelization region is $[-54\text{m}, 54\text{m}]\times[-54\text{m}, 54\text{m}]\times[-5\text{m}, 3\text{m}]$. The model is trained by 20 epochs. Initial learning rate is set to $1e-3$ and then is modified by one-cycle adjuster~\cite{smith2019onecycle}. Adam optimizer~\cite{kingma2014adam} with decay parameters $0.9, 0.99$ is applied. The models are trained by 8 Nvidia V100 GPUs, and the batch size is set to 4 for each GPU. More details are provided in the configuration files of our Github repository~\cite{our-code}.

For KITTI experiments, the resolution for voxelization is set to $[0.16\text{m}, 0.16\text{m}, \text{0.5m}]$ and the voxelization region is $[0\text{m}, 69.12\text{m}]\times[-39.69\text{m}, 39.68\text{m}]\times[-3.5\text{m}, 1.5\text{m}]$. The model is trained by 80 epochs. Initial learning rate is set to $2e-4$ and is decayed by 0.8 for every 15 epochs. Adam optimizer with decay parameters $0.9, 0.99$ is applied. 

Two versions of the \proposenameshort\ are designed in this work: (i) the full version \proposenameshort\ applies RAAConv layers throughout all modules; (ii) \proposenameshort-lite version applies RAAConv layers only at the detection heads. Both designs outperform SOTA baselines. Compared to the \proposenameshort,  \proposenameshort-lite has faster inference speed, with a small drop in performance in terms of the mAP and NDS metric.

\subsection{Training and Inference}
\label{ssec:exp:train_inf}
\proposenameshort\ is trained end-to-end from scratch. Several data augmentation strategies are implemented in the training phase. They are random flipping along the $x$-axis, random scaling for the point clouds, and global rotation around the $z$-axis with a random angle sampled from $[-{\pi}/4,\pi/4]$. 

During inference, we reuse the checkpoint with the best mean average precision (mAP) value and evaluate on the official lidar-only nuScenes and KITTI evaluation metrics. The detection score threshold is set to $0.1$ and IoU threshold is set to $0.2$ for non-maximal suppression. 

The inference speed is tested on the nuScenes. The reason is that the nuScenes data is much closer to the real autonomous driving scenarios, whereas KITTI dataset only stores the front views of the scenes. The latency for the full version of \proposenameshort\ and \proposenameshort-lite are measured as 62 ms and 45 ms per frame, respectively. Both versions satisfy the real-time 3D detection criteria.

\begin{table}[]
\centering
\resizebox{1.0\linewidth}{!}{
\begin{tabular}{l|ccc|c}
\hline
\multicolumn{1}{c|}{Method} & \multicolumn{3}{c|}{BEV (AP$\uparrow$)} &       \\ \hline
                            & Easy  & Moderate & Hard  & AVG   \\ \hline
AVOD\cite{ku2018joint}                       & 0.898 & 0.850    & 0.783 & 0.843 \\
MV3D\cite{chen2017multi}                        & 0.865 & 0.790    & 0.722 & 0.792 \\
PointRCNN\cite{shi2019pointrcnn}                   & 0.921 & 0.874    & 0.827 & 0.874 \\
Fast PointRCNN\cite{chen2019fast}                 & 0.909 & 0.878    & 0.805 & 0.864 \\
F-PointNet\cite{qi2018frustum}                  & 0.912 & 0.847    & 0.748 & 0.835 \\
F-ConvNet\cite{wang2019frustum}                   & 0.915 & 0.858    & 0.761 & 0.845 \\
3DSSD\cite{yang20203dssd}                     & 0.927 & 0.890    & \textbf{0.859} & 0.892 \\
VoxelNet\cite{zhou2018voxelnet}      & 0.880 & 0.784    & 0.713 & 0.792 \\
SECOND\cite{yan2018second}                      & 0.894 & 0.838    & 0.786 & 0.839 \\
PointPillars\cite{lang2019pointpillars}                & 0.901 & 0.866    & 0.828 & 0.865 \\
CenterPoint-pcdet\cite{yin2021center-point}  & 0.885 & 0.851    & 0.812 & 0.849 \\ \hline
RAANet (Ours)               & \textbf{0.953}   & \textbf{0.894} & 0.843 & \textbf{0.897} \\ \hline
\end{tabular}
}
\caption{\small 
\textbf{Comparison with the state-of-the-art models} for BEV detection on the KITTI test dataset. We list Car AP scores for easy, moderate and hard samples. The best results are marked in \textbf{bold}.
}
\label{tab:exp:kitti_results}
\end{table}

\subsection{Results}
\label{ssec:exp:main_results}

\subsubsection{Results on the NuScenes Dataset}
We present the results obtained on the test set with both our lite-version and full-version models, shown in Table~\ref{tab:exp:main_results} and ~\ref{tab:exp:main_results_othermetrics}, in terms of different metrics. The architecture of the full version is illustrated in Fig.~\ref{fig:mtd:flow}. All detection results are evaluated in terms of mean average precision (mAP), mean Average Translation Error (mATE), mean Average Scale Error (mASE), mean Average Orientation Error (mAOE), Average Velocity Error (AVE), Average Attribute Error (AAE) and nuScenes detection score (NDS). The NDS is weighted summed from mAP, mATE, mASE, mAOE, mAVE and mAAE. It firstly converts the errors to scores by $max(1 - TP_error, 0.0)$. Then a weight of 5.0 is assigned to mAP, and a weight of 1.0 is assigned to the other five scores. We compare our work with 16 other SOTA models that have open-source codes. As can be seen in Table~\ref{tab:exp:main_results}, our proposed \proposenameshort\ outperforms all the other models in terms of mAP, and the \proposenameshort-lite takes the second place. For individual classes, \proposenameshort\ achieves the best AP for 9 out of 10 classes, and \proposenameshort-lite takes the second place for 7 out of 10 classes. As shown in Table~\ref{tab:exp:main_results_othermetrics}, the  \proposenameshort\ provides the best performance in terms of mAP, mATE, mAOE, mAVE and NDS metrics, and the 3rd best performance in terms of the mASE and mAAE metrics. The mASE of the RAANet only shows a small gap compared with the top performances on the benchmark. The RAANet-lite takes the second best place for NDS and mAP metrics.

\subsubsection{Results on the KITTI Dataset}
We present the results, obtained on the test set with our full-version models in Table~\ref{tab:exp:kitti_results}. All detection results are evaluated by the average precision (AP) metric for BEV car object detection. We compare our work with other state-of-the-art models that leverage LiDAR points data. As can be seen in Table~\ref{tab:exp:kitti_results}, our proposed \proposenameshort\ outperforms all the other models in terms of the average APs for ``Easy", ``Moderate" and ``Hard" testing samples, which are defined in the KITTI dataset. Thus, when all these testing samples are considered together, \proposenameshort\ achieves the best detection performance.

\begin{figure}[]
\centering
\includegraphics[width=0.4\textwidth]{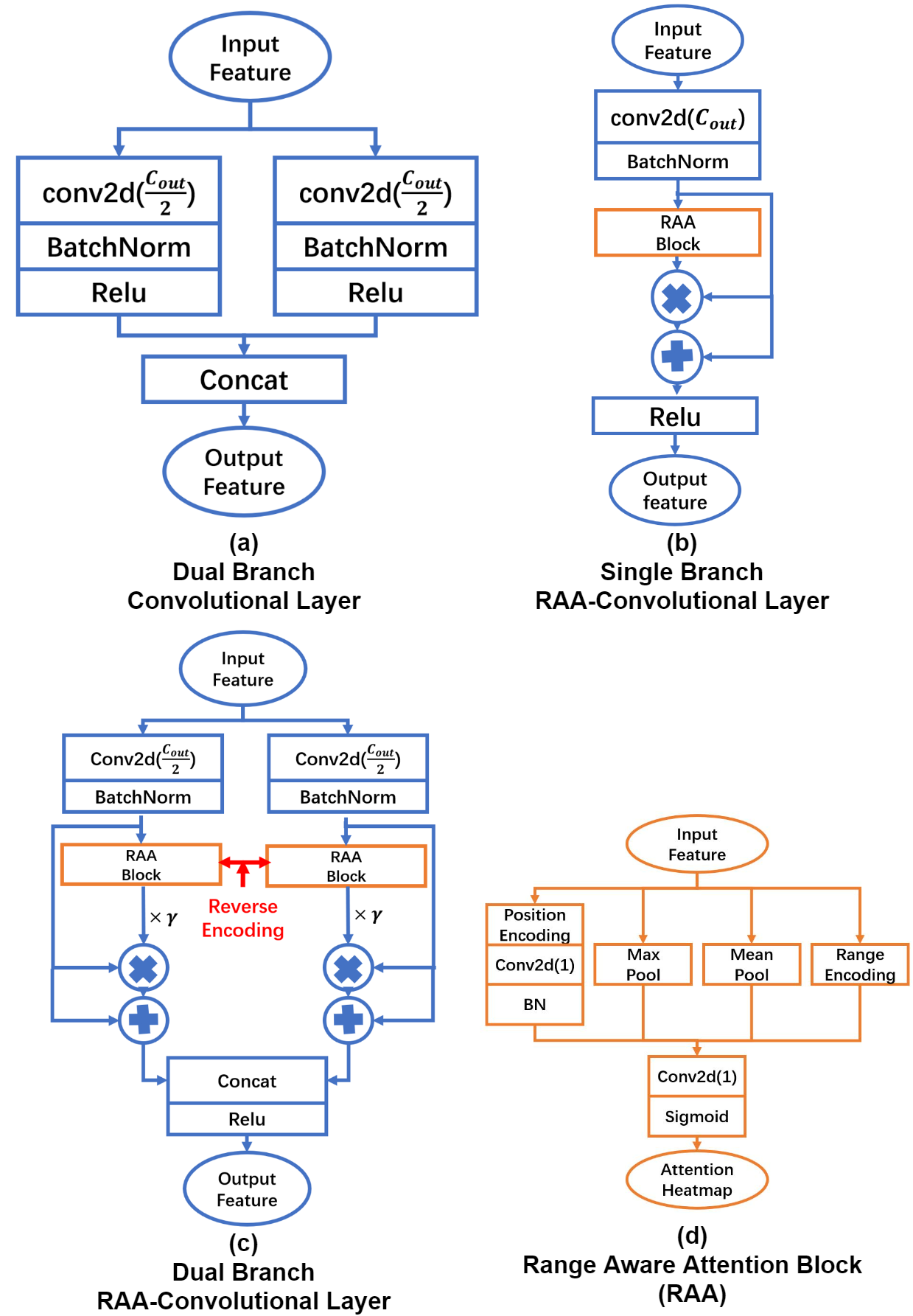}
\caption{\small 
\textbf{RAAConv structures use in the ablation studies.} (a) Dual branch module built with the original convolution, where each branch has half the number of output channels; (b) Single branch module built with range-aware attention, yet without the learnable scalar $\gamma$; (c) Dual branch module built with all sub-components; (d) Structure of the RAA block that is used to generate attention heatmaps.
}
\label{fig:exp:flow}
\end{figure}

\begin{figure}[t!]
\centering
\includegraphics[width=1\columnwidth]{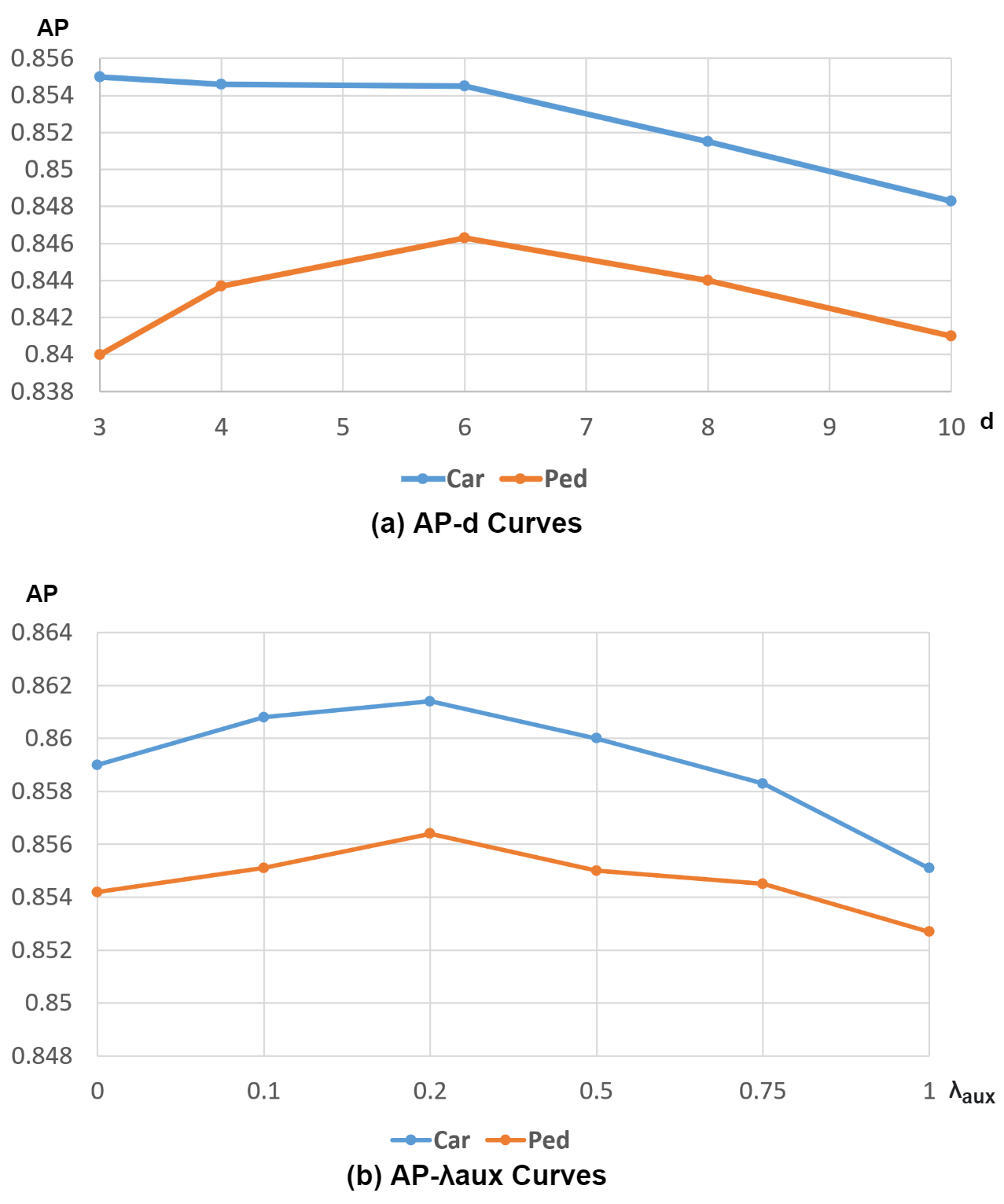}
\caption{\small
\textbf{Ablation Studies on Anistropic Gaussian Mask and ADLE.}
(a) shows AP results with different Gaussian $\Sigma$ decay factors, which is denoted as $d$ in Eq.~(\ref{eq:mtd:gaussian_mask_cont}). (b) shows different ADLE loss weights denoted as $\lambda_{aux}$ in Eq.~(\ref{eq:mtd:overall_loss}).}
\label{fig:exp:sigma_factor}
\end{figure}


\begin{table*}[t!]
\centering
\resizebox{0.9\linewidth}{!}{
\begin{tabular}{l|ccc|c|c|cc|c}
\hline
\multicolumn{1}{c|}{\multirow{2}{*}{Model}} & \multicolumn{3}{c|}{RAAConv}                                                                                                   & \multirow{2}{*}{2D Gaussian} & ADLE & \multicolumn{2}{c|}{AP (\%)} & NDS(\%)  \\ \cline{2-4}
\multicolumn{1}{c|}{}                       & \quad Dual Branch\quad & \quad Attention\quad & $\quad\gamma\quad$ &                              & $l_{aux}=0.25$                 & Car           & Ped.         & All cls. \\ \hline
Baseline                                    & -                                                    & -                                                  & -                  & -                            & -                              & 85.23         & 84.62        & 67.35    \\ \hline
\multirow{4}{*}{+ RAAConv}                  & {\color{green}\cmark}                                                  & {\color{red}\xmark}                                                & {\color{red}\xmark}                & {\color{red}\xmark}                          & {\color{red}\xmark}                            & 85.40         & 84.81        & 68.15    \\
                                            & {\color{red}\xmark}                                                  & {\color{green}\cmark}                                                & {\color{red}\xmark}                & {\color{red}\xmark}                          & {\color{red}\xmark}                            & 85.41         & 84.91        & 68.27    \\
                                            & {\color{green}\cmark}                                                  & {\color{green}\cmark}                                                & {\color{red}\xmark}                & {\color{red}\xmark}                          & {\color{red}\xmark}                            & 85.85         & 85.52        & 68.51    \\
                                            & {\color{green}\cmark}                                                  & {\color{green}\cmark}                                                & {\color{green}\cmark}                & {\color{red}\xmark}                          & {\color{red}\xmark}                            & 86.01         & 85.60        & 69.02    \\ \hline
+ 2D Gaussian                               & {\color{green}\cmark}                                                  & {\color{green}\cmark}                                                & {\color{green}\cmark}                & {\color{green}\cmark}                          & {\color{red}\xmark}                            & 86.25         & -            & 69.59    \\ \hline
+ auxiliary loss                            & {\color{green}\cmark}                                                  & {\color{green}\cmark}                                                & {\color{green}\cmark}                & {\color{green}\cmark}                          & {\color{green}\cmark}                            & 86.41         & 85.72        & 70.27    \\ \hline
\end{tabular}
}
\caption{\small \textbf{Ablation study for the  \proposename.} Various models are built by different combinations of three attributes: RAAConv, Anisotropic Gaussian Mask (2D Gaussian for short) and ADLE. For RAAConv, there are four sub-components: dual branch structure, range-aware attention heatmap and learnable scalar $\gamma$ (Fig.~\ref{fig:mtd:flow}). Anisotropic Gaussian Mask is not applied on the `Pedestrian' class, since the labeled bboxes have similar widths and lengths.
}
\vspace{-0.2cm}
\label{tab:exp:abl_attention}
\end{table*}

\begin{table*}[]
\centering
\resizebox{.75\linewidth}{!}{
\begin{tabular}{c|c|c|c|cc}
\hline
\multirow{2}{*}{\textbf{Models}}\quad         & \multirow{2}{*}{\textbf{\quad RAAConv\quad}} & \multirow{2}{*}{\textbf{\quad 2D Gaussian\quad}} & \multirow{2}{*}{\textbf{\quad ADLE\quad}} & \multicolumn{2}{c}{\textbf{AP(\%)}} \\ \cline{5-6} 
                        &                  &                   &               & Car              & Pedestrian             \\ \hline
Baseline\quad                   & {\color{red}\xmark}               & {\color{red}\xmark}                & {\color{red}\xmark}            & 83.98            & 82.80            \\ \hline
\multirow{4}{*}{RAANet} & {\color{red}\xmark}               & {\color{green}\cmark}               & {\color{red}\xmark}            & 84.32            & -            \\
                        & {\color{green}\cmark}              & {\color{red}\xmark}                & {\color{red}\xmark}            & 84.60            & 83.69            \\
                        & {\color{green}\cmark}              & {\color{green}\cmark}               & {\color{red}\xmark}            & 84.69            & -            \\
                        & {\color{green}\cmark}              & {\color{green}\cmark}               & {\color{green}\cmark}           & 84.80            & 83.79            \\ \hline
\end{tabular}
}
\caption{\textbf{AP results for the \proposenameshort\ with PointPillar Backbone.} The baseline and comparison models are built with a PointPillar backbone and validated on nuScenes. Comparison models are composed by various combinations of RAAConv, 2D Gaussian Mask and ADLE. Similarly, Anisotropic Gaussian Mask is not applied on the ‘Pedestrian’ class.}
\label{tab:exp:ppbackbone}
\end{table*}

\begin{figure*}[h!]
\centering
\includegraphics[width=0.95\textwidth]{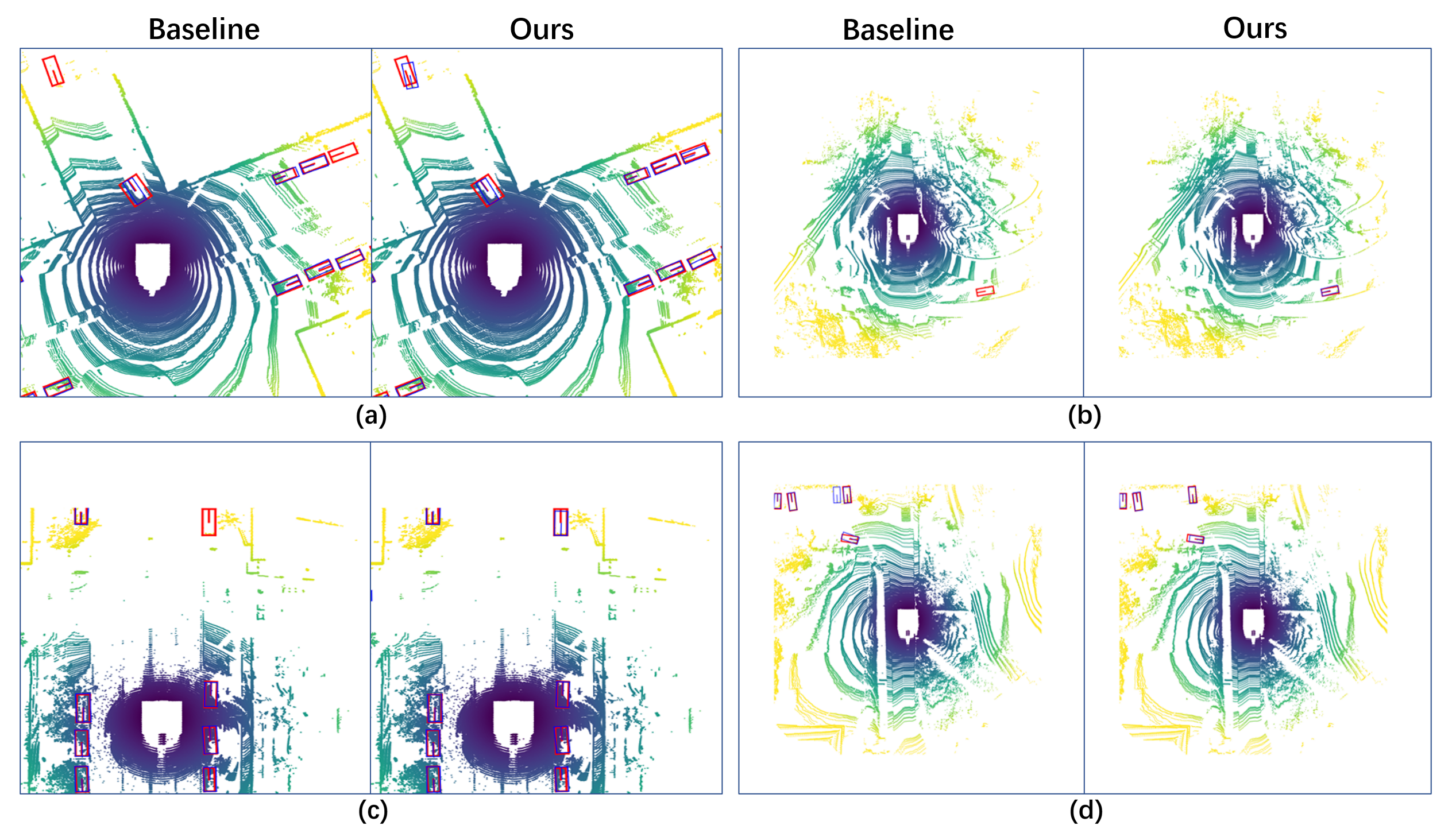}
\caption{\small 
\textbf{Qualitative results.} (Please zoom-in for details). For each image pair, images on the left are the results inferred from the baseline, and the images on the right are the results from the proposed \proposenameshort. {\color{red} Red} and {\color{blue} blue} boxes denote the ground truth and predictions, respectively.
}
\label{fig:exp:compare}
\end{figure*}

\begin{figure*}[]
\centering
\includegraphics[width=0.8\textwidth]{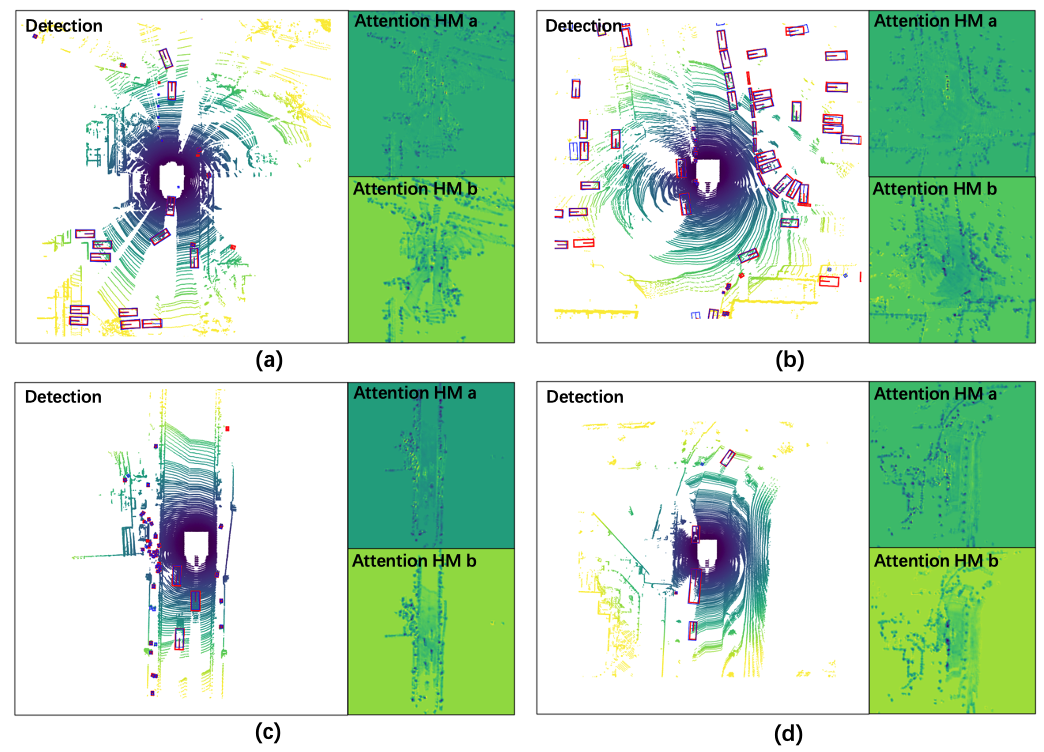}
\caption{\small 
\textbf{Detection and Attention Visualization.} For each sub-figure, the image on the left is the all-class detection result, inferred from our proposed RAANet on the nuScenes validation dataset, and the two images on the right are the attention heatmaps from two convolutional branches in the RAAConv layer. Different color distributions represent how much attentions is paid on the different ranges. 
}
\label{fig:sup:compare}
\end{figure*}

\subsection{Ablation Studies}
\label{sec:exp:ablation}

In this section, we provide the results of the ablation studies for different components and critical parameters of our approach, more specifically the attention module, decay factor $d$ (for the anisotropic Gaussian mask) and $\lambda_{aux}$ (weight for the density level estimation loss). The models are evaluated on the nuScenes validation set, since it has much more samples than the KITTI. For computational efficiency, all the experiments in this section are performed on the `Car' and `Pedestrian' classes to represent large and small objects, respectively. The baseline is the CenterPoint~\cite{yin2021center-point} with Isotropic Gaussian Mask and traditional convolutions.

The structures of the RAAConv layers employed in the ablation studies are shown in Fig.~\ref{fig:exp:flow}. In order to evaluate the effectiveness of our proposed RAAConv layer, we test each sub-component inside the RAAConv. Fig.~\ref{fig:exp:flow} shows three main variants of RAAConv: (a) Dual branch module built with the original convolution, where each branch has half the number of output channels; (b) Single branch module built with range-aware attention, yet without the learnable scalar $\gamma$; (c) Dual branch module built with all sub-components. The reverse encoding indicates different positional encodings on the two attention branches, in order to extract dense and sparse features separately. The last sub-figure (Fig.~\ref{fig:exp:flow}(d)) shows the structure of the RAA block that is used for generating attention heatmaps.

\textbf{Anisotropic Gaussian Mask.} We evaluate the contribution of the Anisotropic Gaussian Mask by varying the decay factor $d$. The ablation study results presented in Fig.~\ref{fig:exp:sigma_factor}(a) show that, with anisotropic Gaussian mask, the performance improvement (in terms of AP) for the `Car' class is more compared to the `Pedestrian' class. One of the probable reasons is that the annotated bboxes for the pedestrian class have similar widths and lengths. This leads the anisotropic Gaussian mask to fall back to the original isotropic Gaussian mask. Additionally, we set $d=3$ for large object and $d=6$ for the small objects in our main experiments. In general, IoU-based evaluation is more sensitive on small objects than large objects, and sharper Gaussian masks help achieve higher accuracy values for small objects.

\textbf{Auxiliary Density Level Estimation Module.} The AP results obtained with various weights of the loss for the auxiliary component ($\lambda_{aux}$), are shown in Fig.~\ref{fig:exp:sigma_factor}(b). The loss weight values are chosen between $0$ and $1$. Based on these results, We set $\lambda_{aux}=0.2$ for the main experiments.

\textbf{Range-Aware Attentional Convolution.} Table~\ref{tab:exp:abl_attention} shows the performance of the baseline as well as several types of the Range-Aware Attention Network.
For RAAConv, there are three sub-components: dual branch structure, range-aware attention heatmap and learnable scalar $\gamma$ (Fig.~\ref{fig:mtd:flow}). Each of these components contributes to the improvement of detection performance. We observe that the combination of dual-branch structure and range-aware attention provides 0.9\% gain on the small object. With the learnable scalar $\gamma$, the dual-branch attentions can be adaptive to the variation of the training data resulting in further improvement.

\begin{figure*}[]
\centering
\includegraphics[width=0.7\textwidth]{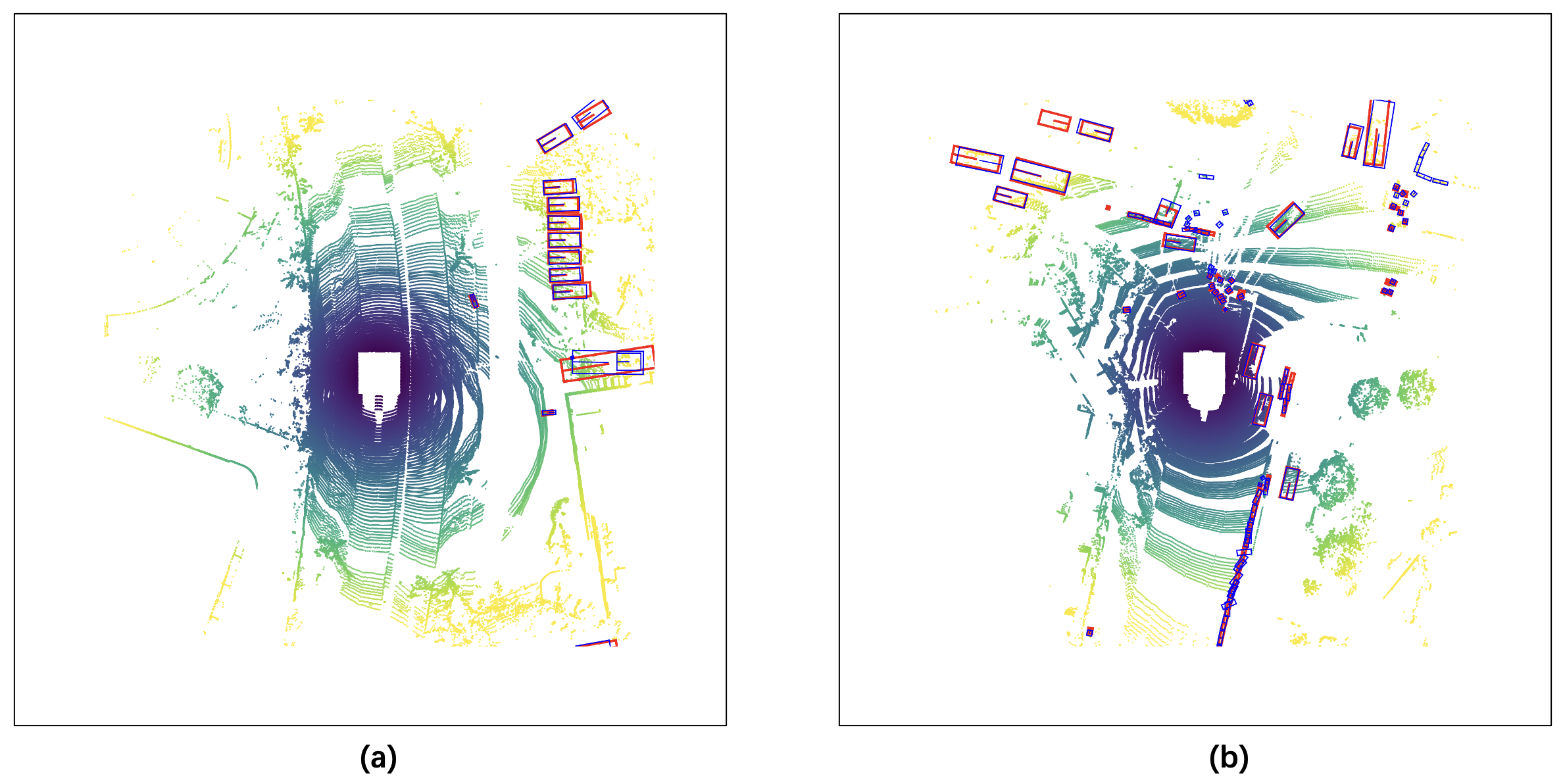}
\caption{\small 
\textbf{Failure Cases.} (a) The far and large objects could cause inaccurate bounding box regression; (b) False positives may appear at the region of sparse point clouds.
}
\label{fig:sup:fail}
\end{figure*}

\subsection{Generalizability}
The proposed \proposenameshort\ is further evaluated by a different backbone from PointPillar~\cite{lang2019pointpillars}. More specifically, we replace the original 3D voxel-based backbone with a 2D pillar-based backbone.~The pillar size is set to $(0.2, 0.2)$. The perception range is set to $[-51.2, 51.2]\times[-51.2, 51.2]\times[-5, 3]$. The baseline is trained with the CenterPoint-PointPillar version. As shown in Table~\ref{tab:exp:ppbackbone}, our proposed \proposenameshort\ still provides a competitive performance with each component contributing to the performance improvement.

\subsection{Qualitative Analysis}
\label{ssec:exp:qualitative}
Pairs of example outputs, obtained by the baseline and the proposed \proposenameshort\ are provided in Fig.~\ref{fig:exp:compare} for qualitative comparison. The sample data is selected from the nuScenes validation dataset. In a pair, images on the left are the results obtained from the baseline, and the images on the right are the results from the proposed \proposenameshort. It can be observed that \proposenameshort\ increases true positives and suppresses false positives at far or occluded regions.

Fig.~\ref{fig:sup:compare} shows more visualization examples on all-class detection of the \proposenameshort\ as well as the attention heatmaps generated from RAAConv. For each sub-figure, the image on the left is the all-class detection result obtained from our proposed RAANet, and the two images on the right are the attention maps from both convolutional branches in the RAAConv layer. It can be observed that both attention modules represent the positional heatmaps based on the BEV. In general, the attention heatmap from branch `a', highlights the regions with dense point clouds or the ego-vehicle vicinity. The attention heatmap from branch `b', on the other hand, pays attention to the regions with sparse point clouds or are farther from the ego vehicle. The combination of both branches effectively extracts powerful BEV features and generates superior 3D object detection results.

Fig.~\ref{fig:sup:fail} shows some hard examples that were misjudged by our proposed method. In Fig.~\ref{fig:sup:fail}(a), although our RAANet successfully generates the detection results for all ground truths, some of the detection headings are not regressed accurately enough due to the long distance and sparse points inside the objects. In Fig.~\ref{fig:sup:fail}(b), some false positives appear on severely occluded regions.

\section{Conclusion}

In this paper, we have introduced the Range-Aware Attention Network (RAANet) to improve the performance of 3D object detection from LiDAR point clouds. The motivation behind the \proposenameshort\ is that, in Bird's Eye View (BEV) LiDAR images, objects appear very different at various distances to the ego-vehicle, and thus, there is a need to avoid using shared-weight convolutional feature extractors. In particular, we have leveraged position and range encodings to design a RAAConv layer with two independent convolution branches, which separately focus on sparse and dense feature extraction. Moreover, we have proposed the auxiliary density level estimation module (ADLE) to further help \proposenameshort\ extract occlusion information of objects during training. Since RAAConv sets the channel number for each branch to the half of final output channels, and ADLE is not attached during inference, the proposed \proposenameshort\ is able to run in real-time. Evaluations on nuScenes and KITTI datasets have shown that our proposed \proposenameshort\ outperforms the SOTA baselines. The code is available at the following link: \href{https://github.com/erbloo/RAAN}{https://github.com/erbloo/RAAN}.

\bibliographystyle{ieee_fullname}
\bibliography{egbib}

\end{document}